\documentclass[10pt,twocolumn,letterpaper]{article}

\usepackage{cvpr}      % To produce the REVIEW version
\usepackage{tcolorbox}
\usepackage{listings}
\usepackage{xcolor}
\lstdefinestyle{promptstyle}{
    basicstyle=\ttfamily\small, % 使用等宽字体并缩小字号
    breaklines=true,            % 自动换行
    breakatwhitespace=true,     % 优先在空格处换行
    columns=fullflexible,       % 保持正常的字符间距
    keepspaces=true,            % 保留代码中的空格缩进
    showstringspaces=false,     % 不显示字符串中的下划线空格
    frame=none                  % 边框由 tcolorbox 提供，这里设为 none
}
\usepackage[accsupp]{axessibility}

\definecolor{cvprblue}{rgb}{0.21,0.49,0.74}
\usepackage[pagebackref,breaklinks,colorlinks,allcolors=cvprblue]{hyperref}

\def\paperID{63} % *** Enter the Paper ID here
\def\confName{CVPR}
\def\confYear{2026}

\title{NTIRE 2026 The 3rd Restore Any Image Model (RAIM) Challenge: \\Professional Image Quality Assessment (Track 1)}

\author{
Guanyi Qin \hspace{1.7em} Jie Liang \hspace{1.7em} Bingbing Zhang \hspace{1.7em} Lishen Qu \hspace{1.7em} Ya-nan Guan \and
Hui Zeng \hspace{1.7em} Lei Zhang \hspace{1.7em} Radu Timofte \hspace{1.7em} Jianhui Sun \hspace{1.7em} Xinli Yue \and
Tao Shao \hspace{1.7em} Huan Hou \hspace{1.7em} Wenjie Liao \hspace{1.7em} Shuhao Han \hspace{1.7em} Jieyu Yuan \and
Chunle Guo \hspace{1.7em} Chongyi Li \hspace{1.7em} Zewen Chen \hspace{1.7em} Yunze Liu \hspace{1.7em} Jian Guo \and
Juan Wang \hspace{1.7em} Yun Zeng \hspace{1.7em} Bing Li \hspace{1.7em} Weiming Hu \hspace{1.7em} Hesong Li \and
Dehua Liu \hspace{1.7em} Xinjie Zhang \hspace{1.7em} Qiang Li \hspace{1.7em} Li Yan \hspace{1.7em} Wei Dong \and
Qingsen Yan \hspace{1.7em} Xingcan Li \hspace{1.7em} Shenglong Zhou \hspace{1.7em} Manjiang Yin \hspace{1.7em} Yinxiang Zhang \and
Hongbo Wang \hspace{1.7em} Jikai Xu \hspace{1.7em} Zhaohui Fan \hspace{1.7em} Dandan Zhu \hspace{1.7em} Wei Sun \and
Weixia Zhang \hspace{1.7em} Kun Zhu \hspace{1.7em} Nana Zhang \hspace{1.7em} Kaiwei Zhang \hspace{1.7em} Qianqian Zhang \and
Zhihan Zhang \hspace{1.7em} William Gordon \hspace{1.7em} Linwei Wu \hspace{1.7em} Jiachen Tu \hspace{1.7em} Guoyi Xu \and
Yaoxin Jiang \hspace{1.7em} Cici Liu \hspace{1.7em} Yaokun Shi
}
\begin{document}
\maketitle
\begin{abstract}
In this paper, we present an overview of the NTIRE 2026 challenge on the 3$^{\text{rd}}$ Restore Any Image Model in the Wild, specifically focusing on Track 1: Professional Image Quality Assessment. Conventional Image Quality Assessment (IQA) typically relies on scalar scores. By compressing complex visual characteristics into a single number, these methods fundamentally struggle to distinguish subtle differences among uniformly high-quality images. Furthermore, they fail to articulate \textit{why} one image is superior, lacking the reasoning capabilities required to provide guidance for vision tasks. To bridge this gap, recent advancements in Multimodal Large Language Models (MLLMs) offer a promising paradigm. Inspired by this potential, our challenge establishes a novel benchmark exploring the ability of MLLMs to mimic human expert cognition in evaluating high-quality image pairs. Participants were tasked with overcoming critical bottlenecks in professional scenarios, centering on two primary objectives: (1) \textbf{Comparative Quality Selection}: reliably identifying the visually superior image within a high-quality pair; and (2) \textbf{Interpretative Reasoning}: generating grounded, expert-level explanations that detail the rationale behind the selection. In total, the challenge attracted nearly 200 registrations and over 2,500 submissions. The top-performing methods significantly advanced the state of the art in professional IQA. The challenge dataset is available \href{https://github.com/narthchin/RAIM-PIQA}{here}, and the official homepage is accessible \href{https://www.codabench.org/competitions/12789/}{here}.
\end{abstract}    
\section{Introduction}
\label{sec:intro}

Image Quality Assessment (IQA) serves as a critical foundation for evaluating visual quality and guiding downstream vision tasks~\cite{qin2023de,qu2026flickerformer,BurstDeflicker_lishenqu}. Despite its fundamental importance, a substantial gap remains between conventional IQA methodologies and the nuanced demands of high-end professional photography. Specifically, such methods typically regress complex visual characteristics into scalar Mean Opinion Scores (MOS) based on low-level statistics or extracted features~\cite{fu2026dr, zhou2025gamma}. However, merely yielding a single number on quality provides limited practical utility. By compressing rich visual information into such a narrow metric, these approaches struggle to distinguish subtle differences among uniformly high-quality images. Furthermore, a solitary score fails to articulate \textit{why} one image is perceived as superior to another, lacking the reasoning capabilities necessary to provide feedback for vision models.

The advent of Multimodal Large Language Models (MLLMs) has established a new pattern for visual quality assessment. Although studies indicate that MLLMs exhibit promising efficacy in quality score regression~\cite{wu2024qalign,zhou2024uniqa}, their actual value should go beyond the number of outputs. MLLMs present a powerful mechanism to overcome the persistent interpretability bottleneck of conventional methods. By coupling robust visual perception with sophisticated linguistic generation, they demonstrate the potential to emulate the cognitive processes of human experts. Specifically, instead of confining complex visual elements (\emph{e.g.}, subtle textures, noise, and artifacts) to latent representations, MLLMs can explicitly leverage these features to formulate grounded reasoning, delivering the nuanced comparative analysis strictly required in professional contexts.

In previous Restore Any Image Model (RAIM) in the Wild challenges, we established a platform for researchers to bridge the gap between academic research and industrial applications, with earlier editions focusing primarily on image enhancement, restoration, and model efficiency. Building upon this solid foundation, the 3$^{\text{rd}}$ RAIM challenge introduces a critical new dimension. Specifically, Track 1 of this year's competition shifts the focus toward Professional Image Quality Assessment. To facilitate this, we provide a comprehensive dataset comprising high-quality image pairs captured in real-world digital photography scenarios. Crucially, these pairs are meticulously annotated by domain experts, who provide not only the ground-truth selection of the visually superior image but also the detailed, text-based reasoning behind their decisions. By leveraging this newly curated dataset, researchers can evaluate the comparative and reasoning capabilities of their MLLMs, validating their models against the high-quality, nuanced feedback of experienced industry practitioners. To foster the development of such MLLM-based algorithms across both academia and industry, the primary objectives of the 3$^{\text{rd}}$ RAIM Track 1 are defined as follows:
\begin{itemize}
    \item Construct a novel benchmark for professional IQA, focusing on \textbf{Comparative Quality Selection} to reliably distinguish the visually superior image within high-quality image pairs.
    \item Promote the development of generic MLLMs capable of \textbf{Interpretative Reasoning}, generating grounded, expert-level explanations that detail the specific reasons for quality variations while avoiding generic or hallucinated descriptions.
    \item Bridge the gap between human expert cognition and automated quality assessment to provide credible, interpretable guidance for real-world photography and downstream image restoration tasks.
\end{itemize}

This challenge is one of the challenges associated with the NTIRE 2026 Workshop~\footnote{  
  \url{https://www.cvlai.net/ntire/2026/} 
  
  \ \ Guanyi Qin, Jie Liang, Bingbing Zhang, Lishen Qu, Ya-nan Guan, Hui Zeng, Lei Zhang, and Radu Timofte are the organizers of the NTIRE 2026 challenge (RAIM Track 1), and other authors are the participants.
  
  \ \ The Appendix lists the authors’ teams and affiliations.  
} on:
deepfake detection~\cite{ntire26deepfake}, 
high-resolution depth~\cite{ntire26hrdepth},
multi-exposure image fusion~\cite{ntire26raim_fusion}, 
AI flash portrait~\cite{ntire26raim_portrait}, 
professional image quality assessment~\cite{ntire26raim_piqa},
light field super-resolution~\cite{ntire26lightsr},
3D content super-resolution~\cite{ntire263dsr},
bitstream-corrupted video restoration~\cite{ntire26videores},
X-AIGC quality assessment~\cite{ntire26XAIGCqa},
shadow removal~\cite{ntire26shadow},
ambient lighting normalization~\cite{ntire26lightnorm},
controllable Bokeh rendering~\cite{ntire26bokeh},
rip current detection and segmentation~\cite{ntire26ripdetseg},
low light image enhancement~\cite{ntire26llie},
high FPS video frame interpolation~\cite{ntire26highfps},
Night-time dehazing~\cite{ntire26nthaze,ntire26nthaze_rep},
learned ISP with unpaired data~\cite{ntire26isp},
short-form UGC video restoration~\cite{ntire26ugcvideo},
raindrop removal for dual-focused images~\cite{ntire26dual_focus},
image super-resolution (x4)~\cite{ntire26srx4},
photography retouching transfer~\cite{ntire26retouching},
mobile real-word super-resolution~\cite{ntire26rwsr},
remote sensing infrared super-resolution~\cite{ntire26rsirsr},
AI-Generated image detection~\cite{ntire26aigendet},
cross-domain few-shot object detection~\cite{ntire26cdfsod},
financial receipt restoration and reasoning~\cite{ntire26finrec},
real-world face restoration~\cite{ntire26faceres},
reflection removal~\cite{ntire26reflection},
anomaly detection of face enhancement~\cite{ntire26anomalydet},
video saliency prediction~\cite{ntire26videosal},
efficient super-resolution~\cite{ntire26effsr},
3d restoration and reconstruction in adverse conditions~\cite{ntire26realx3d},
image denoising~\cite{ntire26denoising},
blind computational aberration correction~\cite{ntire26aberration},
event-based image deblurring~\cite{ntire26eventblurr},
efficient burst HDR and restoration~\cite{ntire26bursthdr},
low-light enhancement: `twilight cowboy'~\cite{ntire26twilight},
and efficient low light image enhancement~\cite{ntire26effllie}.

\section{NTIRE 2026 the 3$^{\text{rd}}$ RAIM Track 1 PIQA}
\label{sec:challenge}

\subsection{Training Data}
\label{sec:training_data}
To support this challenge, we curated a novel reasoning-based dataset format specifically tailored for professional Image Quality Assessment. Each data sample consists of an image pair capturing the same scene, yet acquired using different camera devices or smartphone models, \emph{e.g.}, a specific model versus another. Following the data collection, professional imaging effect designers evaluated each pair. We formulate the identification of the visually superior image as a binary Multiple Choice Question (MCQ), requiring the domain experts to explicitly select either Image A or Image B. To justify their selection, the experts provided comprehensive, text-based reasoning. Furthermore, to facilitate fine-grained observation and explicitly assist the models in their reasoning process, we supplement each image pair with localized cropped patches highlighting the primary subjects. Crucially, the expert annotations delve into professional photographic dimensions, utilizing these localized visual cues to evaluate nuanced attributes such as sharpness, texture details, visual authenticity, and noise characteristics.

\begin{figure*}
    \centering
    \includegraphics[width=\linewidth]{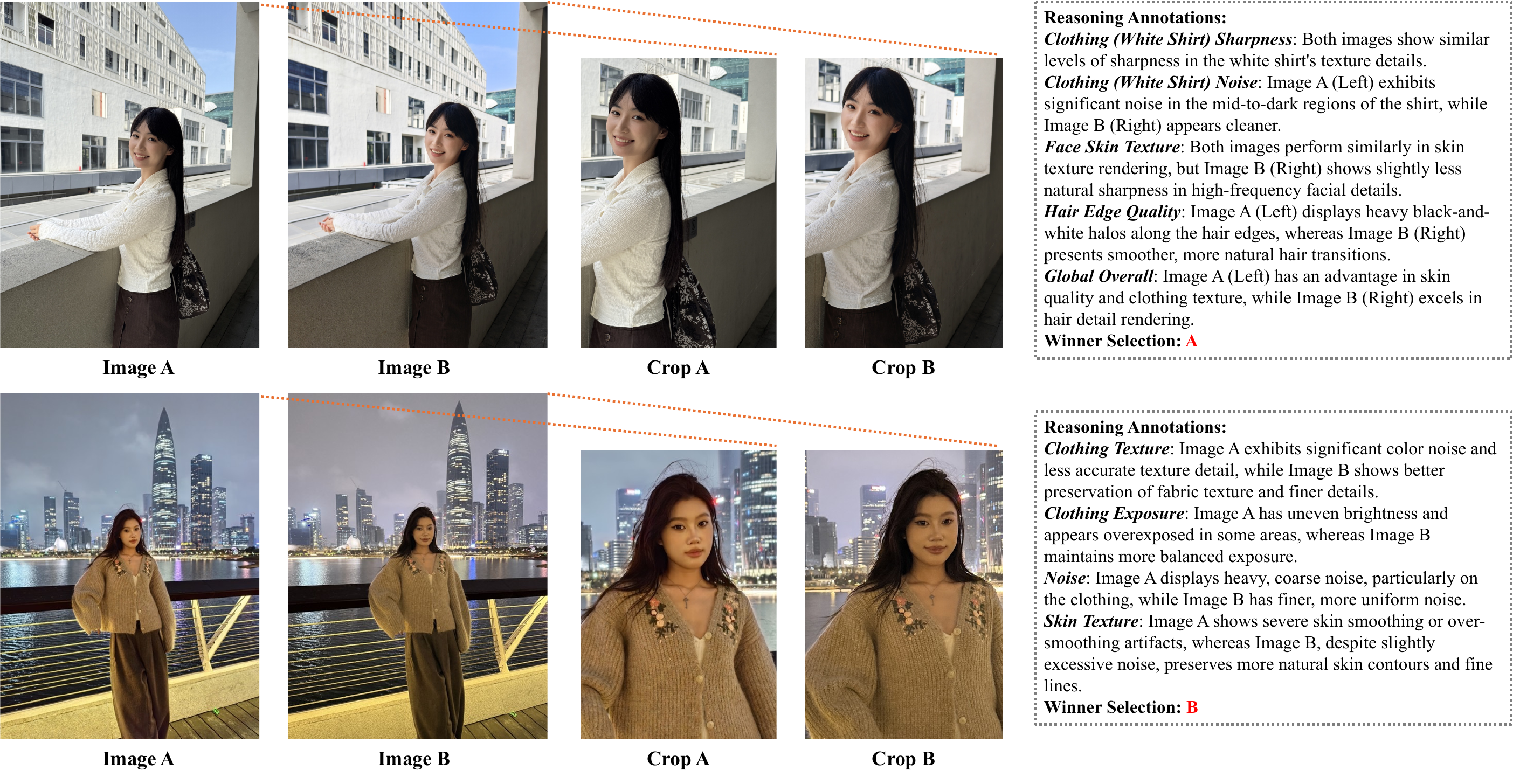}
    \caption{An example of pairwise comparison under the MCQ pattern, with annotations on the rationales behind the selection of the portrait part of the curated dataset.}
    \label{fig:demo_p}
\end{figure*}

\begin{figure*}
    \centering
    \includegraphics[width=\linewidth]{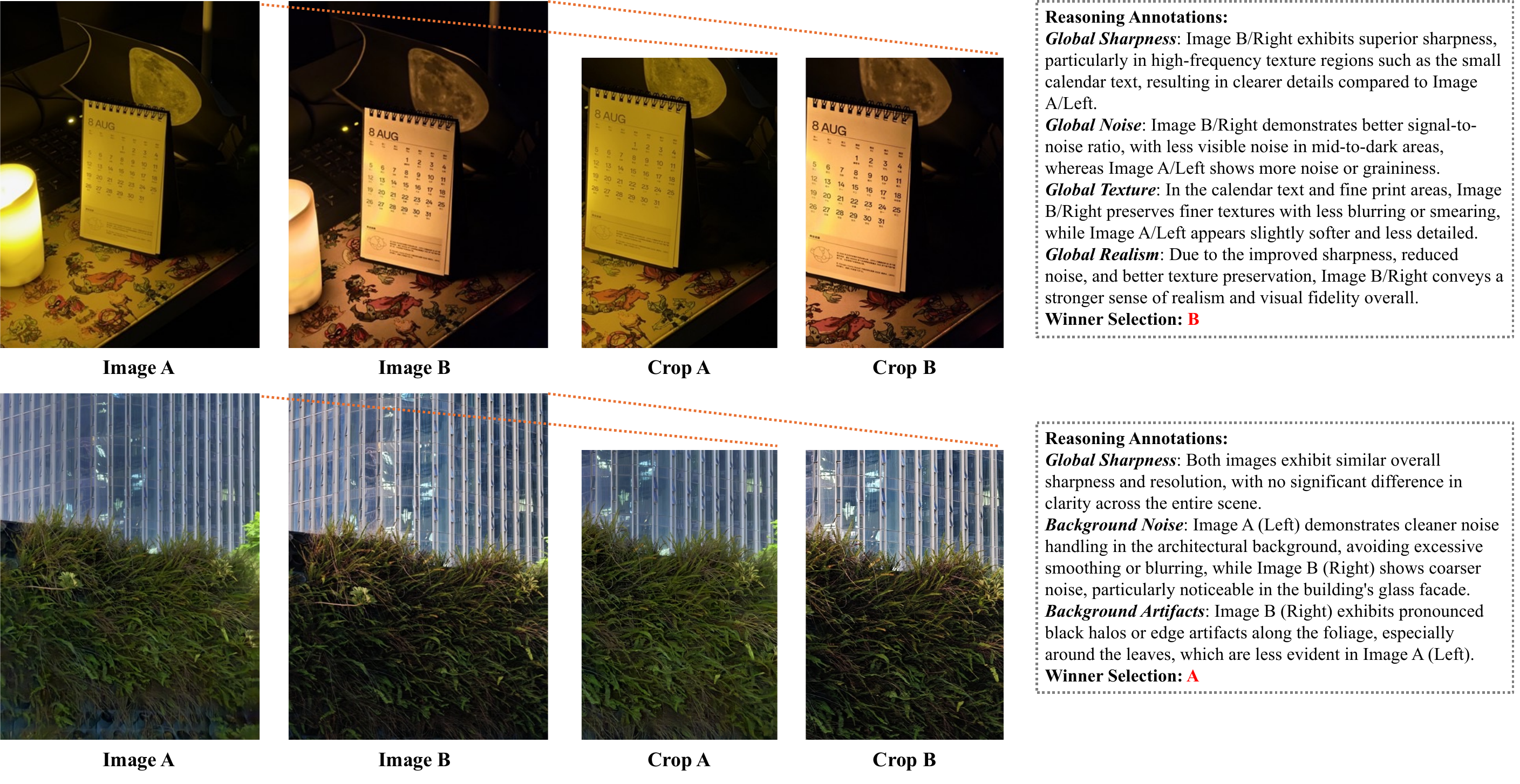}
    \caption{An example of pairwise comparison under the MCQ pattern, with annotations on the rationales behind the selection of the scenery part of the curated dataset.}
    \label{fig:demo_s}
\end{figure*}

In total, the training dataset comprises 100 annotated image pairs. The distribution is specifically designed to reflect common professional photography use cases, containing 75 portrait pairs and 25 landscape and scenery pairs. To ensure the robustness and generalizability of the participating trained models, the dataset exhibits diversity, encompassing a wide array of shooting environments, diverse human subjects, and complex lighting conditions, with examples from portrait pairs and scenery pairs shown in Fig.~\ref{fig:demo_p} \& \ref{fig:demo_s}, respectively. To foster innovation and push the boundaries of professional IQA, this challenge maintains an open-data policy. As in previous editions of RAIM, participants are completely free to leverage any additional data they can curate, alongside any accessible open-source pre-trained models, to develop and optimize their algorithms. However, to ensure fairness and reproducibility, the curation of additional data through manual annotation or the use of closed-source VLMs is strictly prohibited.

\subsection{Validation and Test Data}
\label{sec:val_test_data}
Following the curation pipeline described previously, we compiled a validation set of 102 image pairs, comprising 76 portraits and 26 landscapes, alongside a test set of 101 image pairs, partitioned into 76 portraits and 25 landscapes. To guarantee data diversity, thoroughly evaluate the generalization capabilities of the submitted algorithms, and ensure the practical industrial value of the developed solutions, we meticulously designed the data distribution across these splits. Specifically, the validation set introduces two novel human subjects who are entirely absent from the training data. The test set further incorporates an additional two subjects who remain completely unseen in both the training and validation sets. Regarding the non-portrait data, the landscape and scenery pairs were independently shuffled and sampled from our comprehensive asset library, strictly excluding any scenes previously utilized for training. This rigorous zero-overlap data splitting strategy effectively prevents data leakage and ensures a highly reliable assessment of model performance in real-world, unseen scenarios.

\subsection{Evaluation Measures}
To comprehensively assess the model predictions against the ground truth to reflect performance, we align our evaluation protocol with the objectives of the task. Specifically, the evaluation framework is decomposed into two distinct dimensions:
\begin{itemize}
    \item \textbf{Comparative Accuracy (Acc):} This metric evaluates the model's foundational ability to correctly identify the visually superior image from a given high-quality pair. It serves as the primary quantitative indicator of the model's perceptual judgment.
    \item \textbf{Reasoning Quality:} Beyond mere selection, we rigorously evaluate the interpretability of the models. This dimension measures how closely the generated textual explanations align with the meticulously curated expert annotations, ensuring the reasoning is both professionally grounded and contextually accurate.
\end{itemize}

\subsubsection{Comparative Accuracy}
\label{acc}
To objectively quantify this metric in practice, participating models must structure their output using specific tags, explicitly enclosing their final selection within \texttt{<answer>} and \texttt{</answer>} tags for the MCQ question. The accuracy is then determined through a strict, case-insensitive string match between the model's predicted answer and the ground-truth selection. Ultimately, the final comparative accuracy is calculated as the ratio of correctly predicted pairs to the total number of evaluated image pairs.

\subsubsection{Reasoning Quality Evaluation}
\label{rea}
To systematically assess the reasoning capabilities of the models, we analyze the rationales enclosed within the \texttt{<thinking>} tags. We evaluate this reasoning quality through two complementary approaches: conditional text-based metrics and independent semantic evaluation.

\paragraph{Conditioned NLG Metrics}
To prevent models from being rewarded for generating elaborate but hallucinated justifications for incorrect choices, our traditional text-based evaluation is strictly conditional. \textit{Only} for samples where the model predicts the correct visually superior image, we compute $S_{\text{thinking}}$ as the arithmetic mean of the BLEU-4 score (incorporating method-1 smoothing) and the ROUGE-L F-measure. This composite metric effectively captures both the exact n-gram precision and the structural fluency of the generated rationale.

\paragraph{LLM-as-a-Judge Evaluation}
Recognizing that conventional NLG metrics often struggle to capture deep semantic equivalence and professional nuance, we introduce an auxiliary LLM-as-a-Judge protocol. Independent of the model's comparative selection, an advanced LLM evaluates the semantic alignment between the predicted rationale and the expert ground truth. Specifically, the judge is prompted to verify whether the model's reasoning accurately reflects the advantages of the winning image across designated photographic aspects (\emph{e.g.}, sharpness, texture details, realism, and noise characteristics). This paradigm moves beyond rigid token matching, offering a context-aware assessment to ensure the underlying logic mirrors professional human cognition. The score derived from this evaluation is normalized to a scale of 0 to 1, denoted as $S_{\text{LLM}}$.

\begin{figure*}[t]
\centering
% 配置 tcolorbox 的外观：浅灰背景，圆角，细边框
\begin{tcolorbox}[colback=gray!5, colframe=gray!50, arc=3pt, boxrule=0.5pt, left=4pt, right=4pt, top=4pt, bottom=4pt]
\begin{center}
\textbf{System Prompt for LLM-as-a-Judge Evaluation}
\end{center}
\vspace{-0.5em}
\begin{lstlisting}[style=promptstyle]
You are an expert evaluator for image quality assessment reasoning. Your task is to compare the Candidate Thinking with the Reference (Ground Truth) Thinking.

Please evaluate the candidate strictly based on the following three criteria with their importance order:
1. Aspect Coverage: Does the candidate evaluate the key visual aspects (e.g., face texture, blur, artifacts, noise) mentioned in the reference?
2. Reasoning Correctness: For the matched aspects, is the candidate's visual reasoning and descriptive logic aligned correctly with the reference?
3. Precision & Conciseness: Does the candidate focus on the relevant aspects without excessive verbosity or irrelevant details?

Rate the overall alignment on a scale from 0.0 to 1.0, where:
- 0.0 means completely missing the key aspects or having entirely contradictory reasoning.
- 1.0 means perfectly covering ONLY the reference aspects with entirely correct, aligned, and concise reasoning.
\end{lstlisting}
\end{tcolorbox}
\caption{The structured prompt utilized for the LLM-as-a-Judge protocol to evaluate semantic alignment and reasoning quality.}
\label{fig:prompt}
\end{figure*}

\subsubsection{Ranking Metrics}
\label{ranking}
To establish the leaderboard, we aggregate the aforementioned metrics across different phases of the challenge. For Phase 2, the primary score consolidates basic perceptual accuracy and the conditioned NLG reasoning quality using a multiplicative fusion strategy:
\begin{equation}
    S_{\text{Phase2}} = \text{Accuracy} \times (w_{\text{base}} + w_{\text{reasoning}} \times S_{\text{thinking}})
\end{equation}
where we set $w_{\text{base}} = 0.7$ and $w_{\text{reasoning}} = 0.3$. This weighted formulation ensures that basic accuracy remains the primary driver of the score, while explicitly rewarding high-quality textual reasoning. For Phase 3, we incorporate the semantic evaluation to further distinguish models with superior interpretability. The Phase 3 score $S_{\text{Phase3}}$ builds upon the foundation by adding a scaled LLM judge penalty/bonus:
\begin{equation}
\text{Accuracy} \times (w_{\text{base}} + w_{\text{reasoning}} \times S_{\text{thinking}}) + 0.5 \times S_{\text{LLM}}
\end{equation}
Ultimately, to ensure robustness and consistency, the final ranking score ($S_{\text{Ranking}}$) for the challenge is determined by a balanced ensemble of the performances across both phases:
\begin{equation}
    S_{\text{Ranking}} = 0.5 \times S_{\text{Phase2}} + 0.5 \times S_{\text{Phase3}}
\end{equation}
This multi-tiered evaluation framework comprehensively assesses both the visual judgment and the professional interpretability of the participating models.

\subsection{Phases}

\subsubsection{Phase 1: Model Design and Tuning}
In this initial phase, participants are encouraged to analyze the provided data, design their Multimodal Large Language Model (MLLM) architectures, and tune their algorithms accordingly. As detailed in Section \ref{sec:training_data}, we provided a comprehensive training dataset consisting of 100 annotated image pairs, serving as the fundamental basis for participants to develop and optimize their models' capabilities.

\subsubsection{Phase 2: Online Feedback}
In this phase, participants can submit their predictions to the official evaluation server to receive quantitative feedback. As outlined in Section \ref{sec:val_test_data}, we release the validation set, which includes the image pairs and their corresponding subject crops, while withholding the ground-truth annotations. Participants are required to perform inference on the data using their developed MLLMs. To ensure compatibility with the automated evaluation pipeline, the predictions must be structured containing the image pair identifier, the rationales enclosed within \texttt{<thinking>} tags, and the final selection enclosed within \texttt{<answer>} tags. Upon uploading these formatted results to the server, our deployed scoring program automatically computes the comparative accuracy and the integrated $S_{\text{Phase2}}$. This provides participants with immediate, official feedback on the public leaderboard, allowing them to iteratively refine their algorithms.

\subsubsection{Phase 3: Final Evaluation}
In this culminating phase, we release the test set as detailed in Section \ref{sec:val_test_data}. Consistent with the validation phase, participants are provided solely with the image pairs and corresponding crops, while all annotations are withheld. To ensure a fair and rigorous competitive environment, the submission time window is strictly limited, and each participating team is permitted only a single final submission attempt. Following the closure of the submission portal, the organizing committee conducts an internal verification to reproduce the participants' predictions. Subsequently, the submitted rationales undergo the independent LLM-as-a-Judge semantic evaluation. By integrating these verification steps, the final Phase 3 score ($S_{\text{Phase3}}$) is computed to definitively determine the models' performance.

\subsection{Awards}

The following awards are provided for the Professional IQA track:
\begin{itemize}
\item One first-class award (i.e., the champion) with a cash prize of \textbf{US\$1000};
\item Two second-class awards with cash prizes of \textbf{US\$500 each};
\item Three third-class awards with cash prizes of \textbf{US\$200 each}.
\end{itemize}

\subsection{Important Dates}

\begin{itemize}
\item 2026.01.28: Released data of phase 1. Phase 1 began;
\item 2026.02.08: Released data of phase 2. Phase 2 began;
\item 2026.03.10: Released data of phase 3. Phase 3 began;
\item 2026.03.12: Phase 3 results submission deadline;
\item 2026.03.20: Final rank announced.
\end{itemize}
\section{Challenge Results}

The RAIM Track 1 challenge attracted 192 registered participants. During the Phase 2 validation stage, 85 participants actively contributed a total of 2530 submissions. In the final Phase 3 evaluation, 57 participating teams successfully advanced, resulting in exactly 57 submissions, reflecting the strict single-attempt rule of this phase. The team details and the brief overview of the methods proposed by these teams is detailed in Sections~\ref{sec:teams} and \ref{sec:methods}, respectively.

\begin{table}[!t]
    \centering
    \caption{Award Ranking for the Grand Prize on RAIM Track 1 after organizer-side verification. The final ranking is determined from the scores on Phase 2 and Phase 3.}
    \label{tab:final_results}
    \resizebox{\columnwidth}{!}{
    \begin{tabular}{clccc}
        \toprule
        Rank & Team & Final Award Score & Phase 2 Score & Phase 3 Score \\
        \midrule
        1 & IH-VQA & 0.7305 & 0.6943 & 0.7667 \\
        2 & VCIP Pi Group & 0.7253 & 0.6826 & 0.7679 \\
        3 & I$^2$ Group & 0.7040 & 0.6490 & 0.7590 \\
        4 & fugui & 0.6939 & 0.6640 & 0.7237 \\
        5 & LZ & 0.6823 & 0.6149 & 0.7497 \\
        6 & ongaku & 0.6641 & 0.6056 & 0.7227 \\
        7 & test123djak & 0.6191 & 0.5638 & 0.6743 \\
        8 & dr0strange & 0.6075 & 0.5725 & 0.6424 \\
        9 & NTR & 0.4810 & 0.4418 & 0.5201 \\
        10 & Ayen & 0.3953 & 0.4281 & 0.3626 \\
        \bottomrule
    \end{tabular}}
\end{table}

% ==========================================

\begin{table}[!t]
    \centering
    \caption{Leaderboard of the Phase 3, ranked by Phase 3 Score.}
    \resizebox{\columnwidth}{!}{
    \begin{tabular}{clcccc}
        \toprule
        Rank & Team & Phase 3 Score & LLM Score & Accuracy \\
        \midrule
        1 & VCIP Pi Group & 0.7679 & 0.5535 & 0.6733 \\
        2 & IH-VQA & 0.7667 & 0.4459 & 0.7129 \\
        3 & I$^2$ Group & 0.7590 & 0.4557 & 0.7030 \\
        4 & LZ & 0.7497 & 0.4008 & 0.7228 \\
        5 & fugui & 0.7237 & 0.4137 & 0.6832 \\
        6 & ongaku & 0.7227 & 0.4113 & 0.6832 \\
        7 & test123djak & 0.6743 & 0.3124 & 0.6931 \\
        8 & dr0strange & 0.6424 & 0.3706 & 0.6040 \\
        9 & NTR & 0.5201 & 0.3245 & 0.4752 \\
        10 & Ayen & 0.3626 & 0.1310 & 0.4158 \\
        \bottomrule
    \end{tabular}}
\end{table}

% ==========================================

\begin{table}[!t]
    \centering
    \caption{Leaderboard of the Phase 2, ranked by Phase 2 Score.}
    \resizebox{\columnwidth}{!}{
    \begin{tabular}{clcccc}
        \toprule
        Rank & Team & Phase 2 Score & Accuracy & Conditioned NLG \\
        \midrule
        1 & IH-VQA & 0.6943 & 0.9118 & 0.2050 \\
        2 & VCIP Pi Group & 0.6826 & 0.9314 & 0.1098 \\
        3 & fugui & 0.6640 & 0.8824 & 0.1751 \\
        4 & I$^2$ Group & 0.6490 & 0.8627 & 0.1743 \\
        5 & LZ & 0.6149 & 0.8235 & 0.1557 \\
        6 & ongaku & 0.6056 & 0.8039 & 0.1776 \\
        7 & dr0strange & 0.5725 & 0.7549 & 0.1946 \\
        8 & test123djak & 0.5638 & 0.7549 & 0.1562 \\
        9 & NTR & 0.4418 & 0.5882 & 0.1704 \\
        10 & Ayen & 0.4281 & 0.5980 & 0.0526 \\
        \bottomrule
    \end{tabular}}
\end{table}

\subsection{Quantitative Comparison}
The results as shown in Tab.~\ref{tab:final_results} of results across the three phases demonstrate a highly competitive landscape, with IH-VQA and VCIP Pi Group emerging as the top performers. While IH-VQA secured the first rank in the Final Award Score, VCIP Pi Group showcased superior performance in Phase 3, driven by a notably high LLM Score. The performance gap between the leading teams remained narrow throughout the competition, highlighting the sophisticated optimization strategies employed by the top participants. While some teams excelled in specific metrics, the overall rankings underscore the importance of maintaining balanced capabilities across all evaluation criteria to succeed in this track.

\subsection{Qualitative Comparison}
\begin{figure*}
    \centering
    \includegraphics[width=\linewidth]{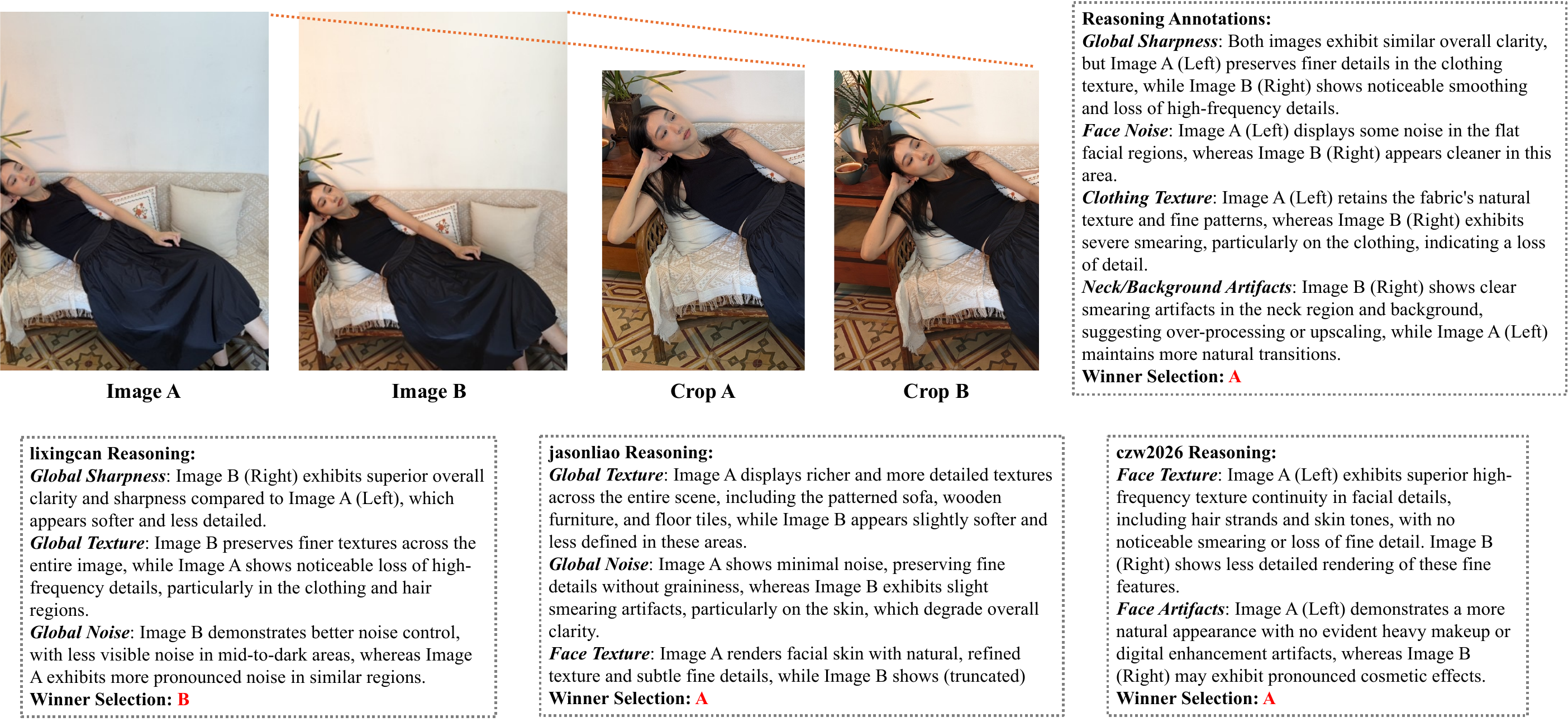}
    \caption{Visual comparison of generated rationales and selections with ground-truth annotations and selections on the portrait part of the curated test set.}
    \label{fig:compar_p}
\end{figure*}

\begin{figure*}
    \centering
    \includegraphics[width=\linewidth]{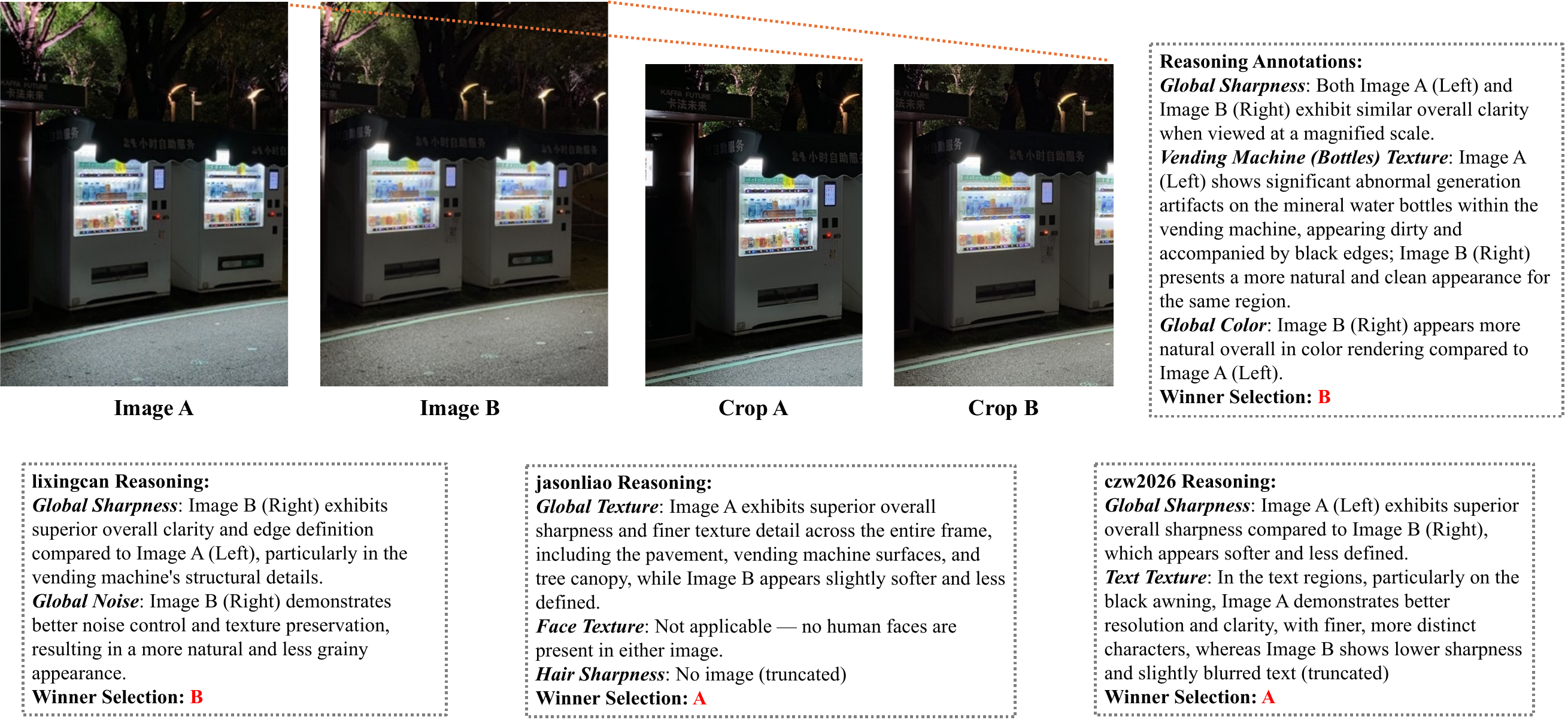}
    \caption{Visual comparison of generated rationales and selections with ground-truth annotations and selections on the scenery part of the curated test set.}
    \label{fig:compar_s}
\end{figure*}

Figs.~\ref{fig:compar_p} and \ref{fig:compar_s} provide representative visual comparisons between the reasoning rationales generated by participating models and the curated ground-truth annotations from our test set. As illustrated, while the models generally demonstrate proficiency in making binary selections, generating high-quality reasoning remains a significant challenge. Specifically, many models struggle to produce rationales that align with the nuanced photographic dimensions identified by experts. Furthermore, evaluating these reasoning outputs across diverse image contents and shooting conditions remains an open problem, as the optimal criteria for assessment can vary significantly depending on the specific visual context and artifacts present in each pair.

\section{Teams \& Affiliations}
\label{appendix}

\textbf{NTIRE 2026 Team}
\label{sec:teams}
{\flushleft

\noindent\textit{\textbf{Challenge:}} 

\noindent NTIRE 2025 Restore Any Image Model (RAIM) in the Wild Track 1: Professional Image Quality Assessment

\noindent\textit{\textbf{Organizers:}}

\noindent Guanyi Qin$^{1,2}$ (qinguanyi@oppo.com)

\noindent Jie Liang$^1$ (liang27jie@gmail.com)

\noindent Bingbing Zhang$^1$ (haiyang19@gmail.com)

\noindent Lishen Qu$^{1,3}$ (qulishen@mail.nankai.edu.cn)

\noindent Ya-nan Guan$^{1,3}$ (guanyanan@mail.nankai.edu.cn)

\noindent Hui Zeng$^{1}$ (cshzeng@gmail.com)

\noindent Prof. Lei Zhang$^{1,4}$ (cslzhang@comp.polyu.edu.hk)

\noindent Prof. Radu Timofte$^5$ (radu.timofte@uni-wuerzburg.de)

\noindent\textit{\textbf{Affiliations:}}

\noindent $^1$ OPPO Research Institute, China

\noindent $^2$ National University of Singapore, Singapore

\noindent $^3$ Nankai University, China

\noindent $^4$ The Hong Kong Polytechnic University, China

\noindent $^5$ Computer Vision Lab, University of W\"urzburg, Germany

~\\

\textbf{Participating Team Details}

\noindent \textit{\textbf{Team name:}} IH-VQA~\cite{ntire26idff}\\
\textit{\textbf{Members:}}\\
Jianhui Sun$^1$ (nimosun@tencent.com), Xinli Yue$^1$, Tao Shao$^1$, Huan Hou$^1$\\
\textit{\textbf{Affiliations:}}\\
$^1$ WeChat, Tencent, China

~\\

\noindent \textit{\textbf{Team name:}} VCIP Pi Group~\cite{ntire26vlmiqa}\\
\textit{\textbf{Members:}}\\
Wenjie Liao$^1$ (2120250634@mail.nankai.edu.cn), Shuhao Han$^1$, Jieyu Yuan$^1$, Chunle Guo$^1$, Chongyi Li$^1$\\
\textit{\textbf{Affiliations:}}\\
$^1$ Nankai University

~\\

\noindent \textit{\textbf{Team name:}} I$^2$ Group\\
\textit{\textbf{Members:}}\\
Zewen Chen$^{1,2}$ (chenzewen2022@ia.ac.cn), Yunze Liu$^3$, Jian Guo$^4$, Juan Wang$^1$, Yun Zeng$^5$, Bing Li$^{1,6}$, Weiming Hu$^{1,2,7}$, Hesong Li$^8$, Dehua Liu$^8$\\
\textit{\textbf{Affiliations:}}\\
$^1$ State Key Laboratory of Multimodal Artificial Intelligence Systems, CASIA\\
$^2$ School of Artificial Intelligence, University of Chinese Academy of Sciences\\
$^3$ Zhongguancun Academy, Beijing 100094, China\\
$^4$ Beijing Union University\\
$^5$ China University of Petroleum\\
$^6$ PeopleAl Inc., Beijing, China\\
$^7$ School of Information Science and Technology, ShanghaiTech University\\
$^8$ ChuanYin, China

~\\

\noindent \textit{\textbf{Team name:}} fugui\\
\textit{\textbf{Members:}}\\
Xinjie Zhang$^1$ (1048509754@qq.com), Qiang Li$^1$, Li Yan$^1$, Wei Dong$^2$, Qingsen Yan$^1$\\
\textit{\textbf{Affiliations:}}\\
$^1$ Northwestern Polytechnical University, China\\
$^2$ Xi'an University of Architecture and Technology, China

~\\

\noindent \textit{\textbf{Team name:}} LZ\\
\textit{\textbf{Members:}}\\
Xingcan Li$^1$ (13308340381@163.com), Shenglong Zhou$^2$, Manjiang Yin$^2$\\
\textit{\textbf{Affiliations:}}\\
$^1$ Chongqing University, China\\
$^2$ University of Science and Technology of China, China

~\\

\noindent \textit{\textbf{Team name:}} ongaku\\
\textit{\textbf{Members:}}\\
Yinxiang Zhang$^1$ (zhangyinxiang@mail.nwpu.edu.cn), Hongbo Wang$^1$\\
\textit{\textbf{Affiliations:}}\\
$^1$ Northwestern Polytechnical University, China

~\\

\noindent \textit{\textbf{Team name:}} test123djak\\
\textit{\textbf{Members:}}\\
Jikai Xu$^1$ (51285901144@stu.ecnu.edu.cn), Zhaohui Fan$^1$, Dandan Zhu$^1$, Wei Sun$^1$, Weixia Zhang$^2$, Kun Zhu$^3$, Nana Zhang$^4$, Kaiwei Zhang$^5$\\
\textit{\textbf{Affiliations:}}\\
$^1$ School of Computer Science and Technology, East China Normal University, China\\
$^2$ School of Computer Science, Shanghai Jiao Tong University, China\\
$^3$ School of Computer Science, Tongji University, China\\
$^4$ School of Information and Intelligent Science, Donghua University, China\\
$^5$ Shanghai Artificial Intelligence Laboratory

~\\

\noindent \textit{\textbf{Team name:}} dr0strange\\
\textit{\textbf{Members:}}\\
Qianqian Zhang$^1$ (qiz087@ucsd.edu), Zhihan Zhang$^1$, William Gordon$^1$, Linwei Wu$^1$\\
\textit{\textbf{Affiliations:}}\\
$^1$ University of California, San Diego, United States

~\\

\noindent \textit{\textbf{Team name:}} NTR\\
\textit{\textbf{Members:}}\\
Jiachen Tu$^1$ (jtu9@illinois.edu), Guoyi Xu$^1$, Yaoxin Jiang$^1$, Cici Liu$^1$, Yaokun Shi$^1$\\
\textit{\textbf{Affiliations:}}\\
$^1$ University of Illinois Urbana-Champaign

~\\

}

\subsection{Submitted Methods}
\label{sec:methodso}
A key trend in this competition was the widespread adoption of, specifically leveraging Group Relative Policy Optimization (GRPO), alongside voting ensembles as strategies. Due to space limitations, the methods proposed by the participating teams are described in the material.
\section{Conclusion}
In this paper, we summarized Track 1 of the NTIRE 2026 3$^{\text{rd}}$ RAIM in the Wild challenge: Professional Image Quality Assessment. To address the limitations of conventional scalar-based IQA, this challenge championed a paradigm shift toward Multimodal Large Language Models (MLLMs). By introducing a meticulously curated benchmark of high-quality image pairs with expert-annotated reasoning, we established a rigorous testing ground far beyond simple numerical regression. Overall, the innovative solutions from competing teams clearly underscore the viability of MLLMs in bridging automated assessment and human cognition. Top-performing methodologies demonstrated both robust comparative quality selection and remarkable interpretative reasoning, yielding grounded feedback. Ultimately, we hope our datasets and multi-tiered metrics will inspire the community, transitioning the field from scoring toward industrially valuable evaluation systems for next-generation photography and vision.

\section*{Acknowledgments}
This work was partially supported by the Humboldt Foundation. We thank the NTIRE 2026 sponsors: OPPO, Kuaishou, and the University of Wurzburg (Computer Vision Lab).
{
    \small
    \bibliographystyle{ieeenat_fullname}
    \bibliography{main}
}

\clearpage
\appendix
\section{Team \& Methods}
\label{sec:methods}

This section briefly describes the participating teams and their proposed methods for this track.

\subsection{IH-VQA - Methodology}
\label{sec:ih-vqa}
The team's solution adopts a dual-branch framework for pairwise image quality assessment in professional photography scenarios. Specifically, the team designs an Answer Model for robust binary preference prediction and a Thinking Model for expert-style rationale generation. The former focuses on accurate decision making, while the latter improves interpretability by producing structured textual explanations for the predicted preference.

\subsubsection{Data Synthesis and Pre-processing} 
Each sample contains a pair of compared images with both global and local cropped views. For the Answer Model, instead of directly using the original concatenated pair image, the team explicitly splits each sample into four inputs: global-left, global-right, crop-left, and crop-right. This formulation allows the model to better capture relative quality differences between the two candidates. In addition, the researchers divide the data into two semantic subsets, namely person and scene, and train separate models for each subset to reduce content heterogeneity. For the Thinking Model, the team progressively enriches the instruction data by introducing expert-style templates, traditional CV features, learned IQA features, and answer-aware prompts.

\subsubsection{Network Design} 
The Answer Model is formulated as a discriminative pairwise classifier. The four decomposed views are processed by a shared visual backbone, and the extracted features are fused for left-right preference prediction. To improve generalization, the team explores multiple backbone architectures and combines their predictions through an ensemble approach. In addition, separate person and scene branches are trained to better model domain-specific quality cues.

\begin{figure}
    \centering
    \includegraphics[width=\linewidth]{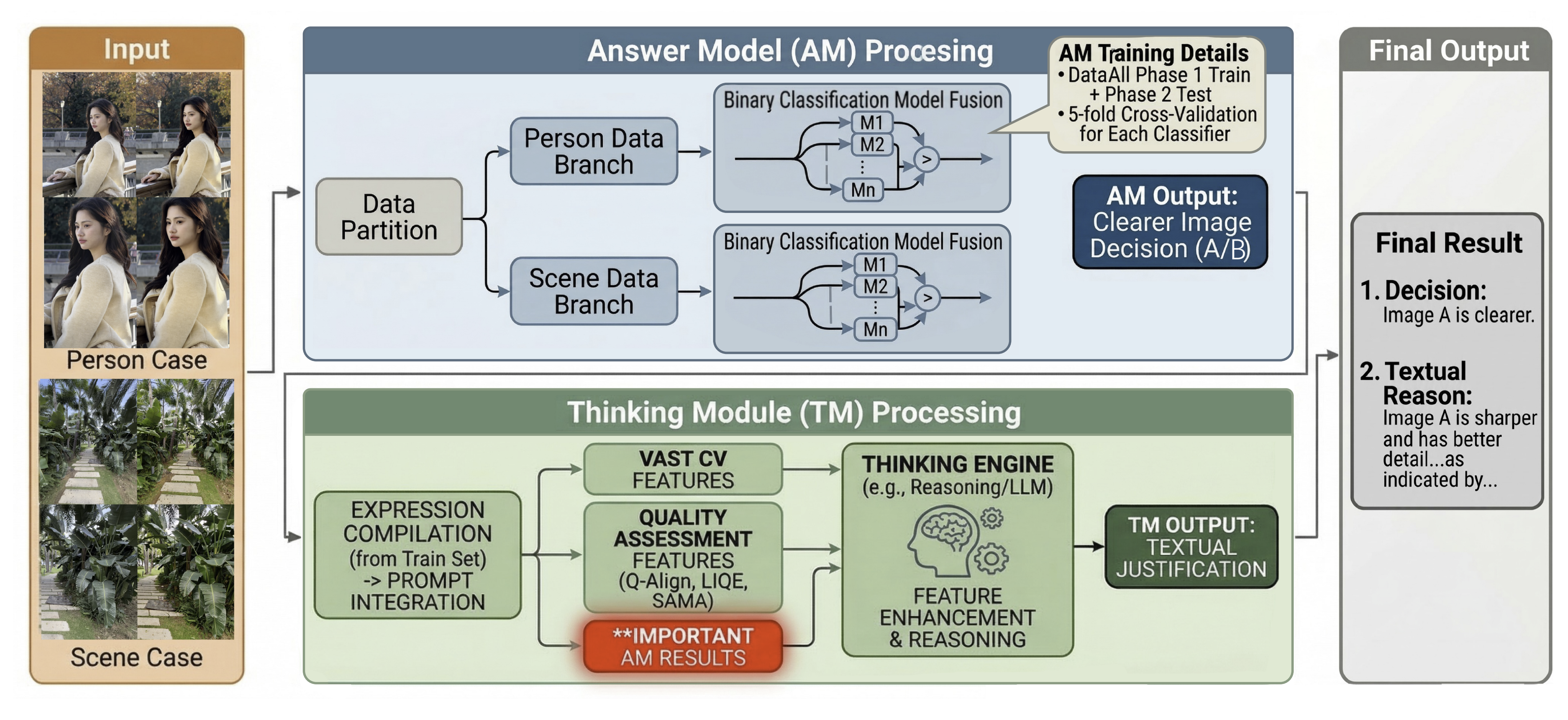}
    \caption{Overview of the proposed dualbranch framework of IH-VQA for pairwise image quality assessment. The Answer Model (AM) performs robust preference prediction through contentaware branching and model fusion, while the Thinking Model (TM) generates expertstyle textual rationales by integrating prompt templates, qualityrelated features, and Answer Model outputs. The final output includes both the preferred image decision and its corresponding explanation.}
    \label{fig:ihvqa}
\end{figure}

The Thinking Model is built on a multimodal large language model. Given the image pair and the structured prompt, it generates expert-style reasoning for the quality comparison task. The team progressively enhances the prompting scheme with domain-specific templates, traditional CV statistics, learned IQA metrics (e.g., Q-Align, LIQE, and SAMA), and answer-aware rationale refinement, which substantially improves the quality and consistency of the generated explanations.

\subsubsection{Training Details} 
For the Answer Model, the team trains binary classifiers with AdamW for $10$ epochs using an initial learning rate of $1 \times 10^{-4}$, weight decay of $1 \times 10^{-4}$, and a batch size of $32$. Images are resized to $256 \times 256$ and randomly cropped to $224 \times 224$. The researchers adopt a $5$-fold training strategy and use fold-wise voting together with a multi-backbone ensemble for robust final prediction.

For the Thinking Model, the team fine-tunes \textit{Qwen3-VL-8B-Instruct} with LoRA-based supervised fine-tuning. The training is conducted for $6$ epochs with a learning rate of $4 \times 10^{-5}$, LoRA rank $r=16$, and DeepSpeed ZeRO-1 on $8$ GPUs.

\subsection{VCIP Pi Group - Methodology}
\label{sec:vcip-pi-group}
The VCIP Pi Group proposes a single-agent framework based on vision-language models (VLMs) for fine-grained, paired image quality comparison across nine expert-defined dimensions. The architecture utilizes a two-tier design based on the \textit{Qwen3-VL} series: nine \textit{Qwen3-VL-2B-Instruct} models serve as dimension-specific tools, while one \textit{Qwen3-VL-8B-Instruct} model acts as the core agent. This core agent employs a closed-loop Planner-Executor-Summarizer pipeline with memory updates to process global and local image pairs for comprehensive decision-making.

\subsubsection{Data Synthesis and Pre-processing}
The data construction pipeline is a two-stage, LLM-driven framework. In the first stage, multiple LLMs generate expert-level analysis text across nine predefined dimensions for both global and local image regions. In the second stage, ground-truth tool calls mapping each analysis to a specific \texttt{\{Tool, Region\}} pair are extracted. The data is further augmented through local image pair swapping and refined using an LLM judge to filter out low-quality reasoning data and ensure reliable annotations.

\begin{figure}
    \centering
    \includegraphics[width=\linewidth]{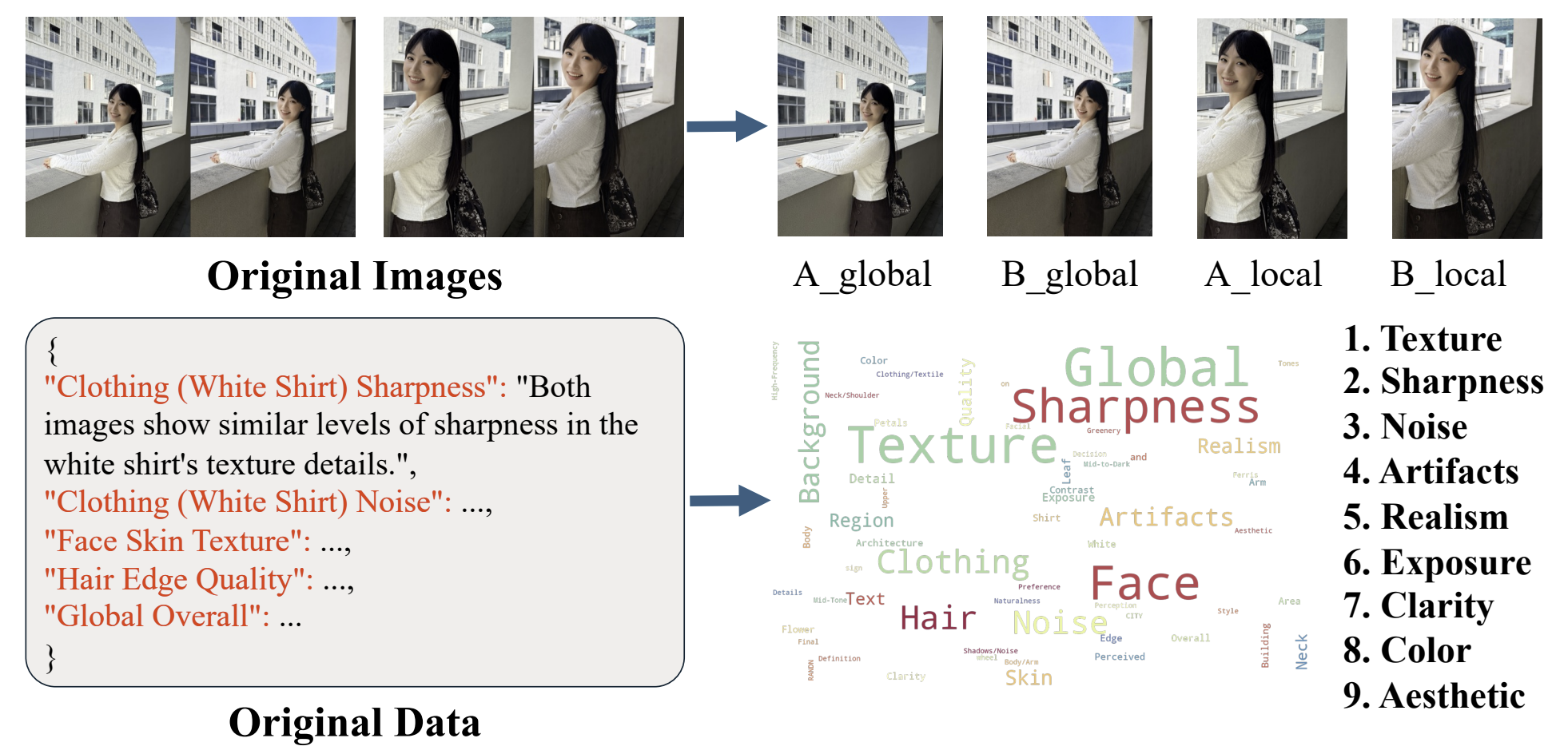}
    \caption{Data construction pipeline of VCIP Group.}
    \label{fig:vcip-data}
\end{figure}

\subsubsection{Network Design}
The framework is built upon the \textit{Qwen3-VL} backbone, embedding task-specific modules into its reasoning layer without altering the core structure. The core \textit{Qwen3-VL-8B-Instruct} agent coordinates the nine lightweight \textit{Qwen3-VL-2B-Instruct} dimension tools via a unified memory bank and a cyclic reasoning pipeline:
\begin{itemize}
    \item \textbf{Planner Module:} Parses the input image pairs, identifies key regions, generates an initial \texttt{\{Tool, Region\}} call sequence, and initializes the memory bank.
    \item \textbf{Executor Module:} Dispatches the corresponding 2B dimension tools, updates the memory bank with their structured outputs, and determines whether to continue the loop or terminate based on maximum iteration limits or duplicate call detection.
    \item \textbf{Summarizer Module:} Extracts all single-dimension results from memory to perform comprehensive multi-dimensional reasoning, outputting the final quality comparison conclusion in a standardized format.
\end{itemize}

\begin{figure}[!ht]
    \centering
    \includegraphics[width=\linewidth]{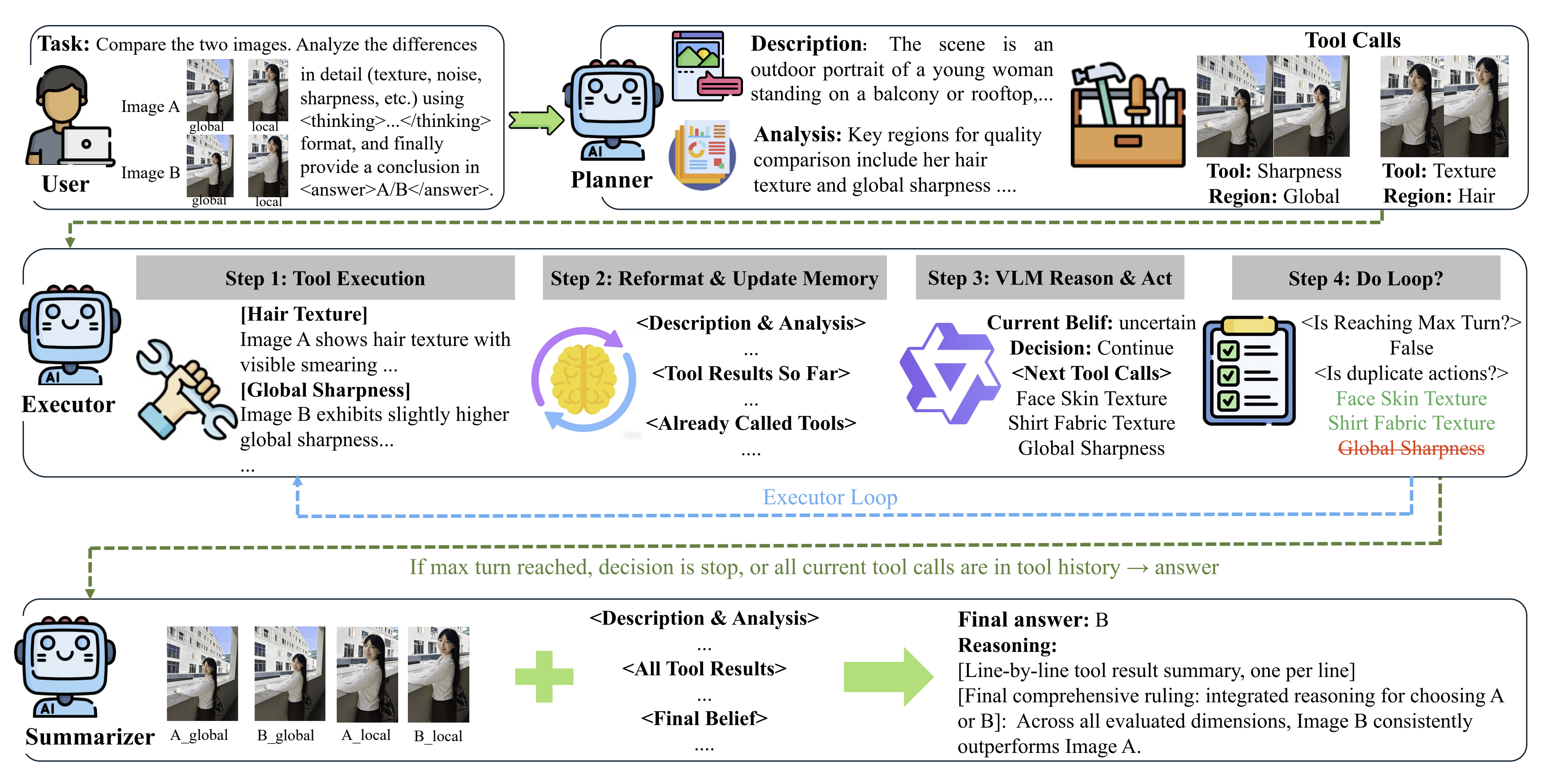}
    \caption{Agent framework of VCIP Group.}
    \label{fig:vcip-agent}
\end{figure}

\subsubsection{Training Details}
The training framework adopts a lightweight LoRA fine-tuning paradigm applied to the Vision Encoder and the \textit{Qwen3} LM Dense Decoder, while keeping the Text Tokenizer frozen to maintain consistent text encoding. The team utilizes differentiated training pipelines for the two tiers:
\begin{itemize}
    \item \textbf{Tool Layer (2B models):} Independently undergoes a two-stage sequential workflow (SFT followed by GRPO) under ground-truth supervision to learn basic visual feature analysis and structured output generation.
    \item \textbf{Core Agent Layer (8B model):} Trained solely via GRPO to optimize tool scheduling, reasoning, and result fusion. Hardware allocation includes $8$ NVIDIA H20 GPUs for large-scale agent training and $9$ NVIDIA L40 GPUs to support parallel tool inference.
\end{itemize}

\begin{figure*}[!ht]
    \centering
    \includegraphics[width=\linewidth]{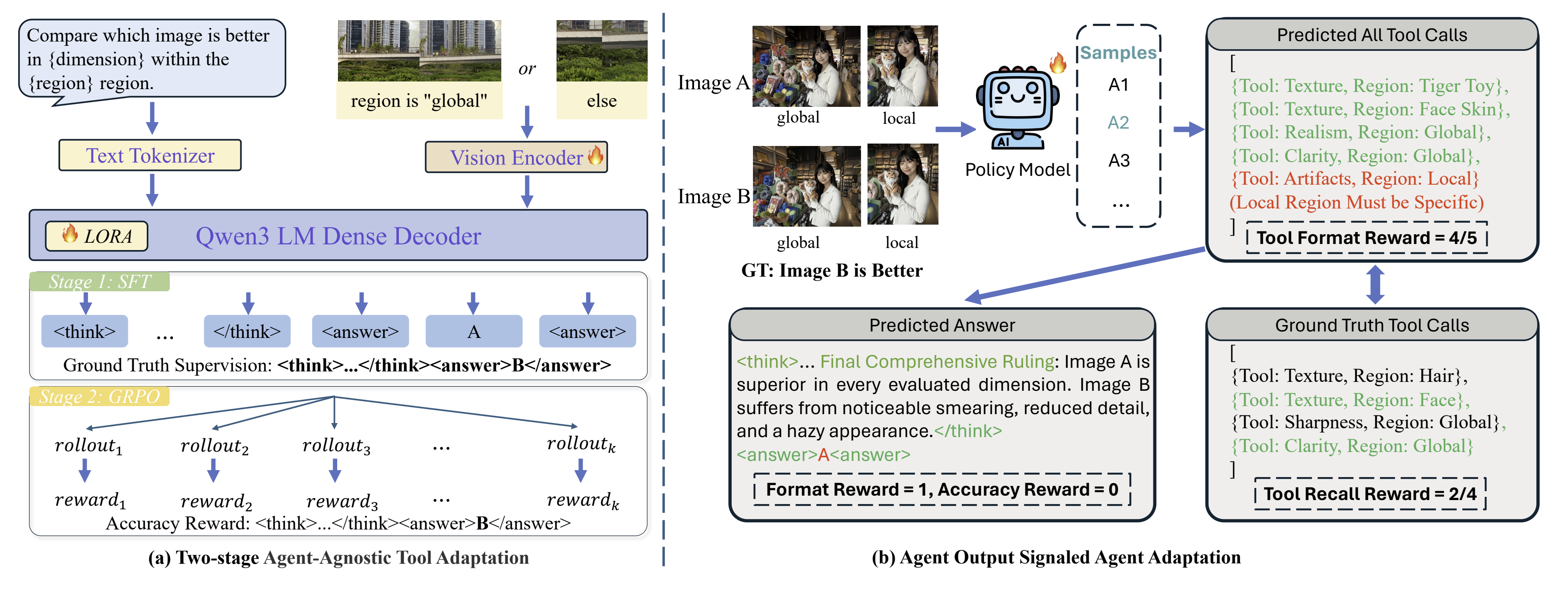}
    \caption{Training pipeline of VCIP Group}
    \label{fig:vcip-train}
\end{figure*}

To guide the GRPO training, the researchers implement a customized multi-dimensional reward mechanism applied to the agent's output trajectories. This includes a \textbf{Format Reward} for structural compliance (e.g., correct reasoning and answer tags), an \textbf{Accuracy Reward} for verifying final answer consistency against the ground truth, a \textbf{Tool Format Reward} to enforce valid JSON formats and penalize duplicate calls, and a \textbf{Tool Recall Reward} to measure the alignment between predicted and ground-truth tool combinations.

\subsection{I$^2$ Group - Methodology}
\label{sec:i2-group}
The I$^2$ Group proposes a multi-stage training and ensemble framework built on top of \textit{Qwen3-VL-8B-Instruct} for pairwise professional image quality assessment. The backbone is progressively optimized with alternating supervised fine-tuning (SFT) and GRPO-based post-training. The final prediction is obtained by majority voting over three complementary expert checkpoints. This design improves both reasoning consistency and decision robustness on comparison-based IQA samples.

\subsubsection{Data Synthesis and Pre-processing}
The team does not synthesize extra images. Instead, they reformulate the official data by splitting each merged detailed image into two single-view crops corresponding to Image A and Image B. All training and inference samples are built on these split-zoom images. Based on the official annotations, the researchers further construct four JSONL datasets: an original SFT set, an instruction-expanded SFT set with more question styles, an original GRPO set, and a swapped-order GRPO set that exchanges the two inputs and updates the corresponding ground-truth answer.

\subsubsection{Network Design}
The backbone is \textit{Qwen3-VL-8B-Instruct} implemented with the \textit{ms-swift} framework. The overall training pipeline contains five stages: SFT, GRPO, GRPO, SFT, and final GRPO. In GRPO, the team uses two reward terms, including answer correctness and output-format compliance with the required thinking and answer structure. For the final system, they select three expert models from Stage $1$, Stage $2$, and Stage $5$, respectively, and ensemble them with majority voting.

\begin{figure}
    \centering
    \includegraphics[width=\linewidth]{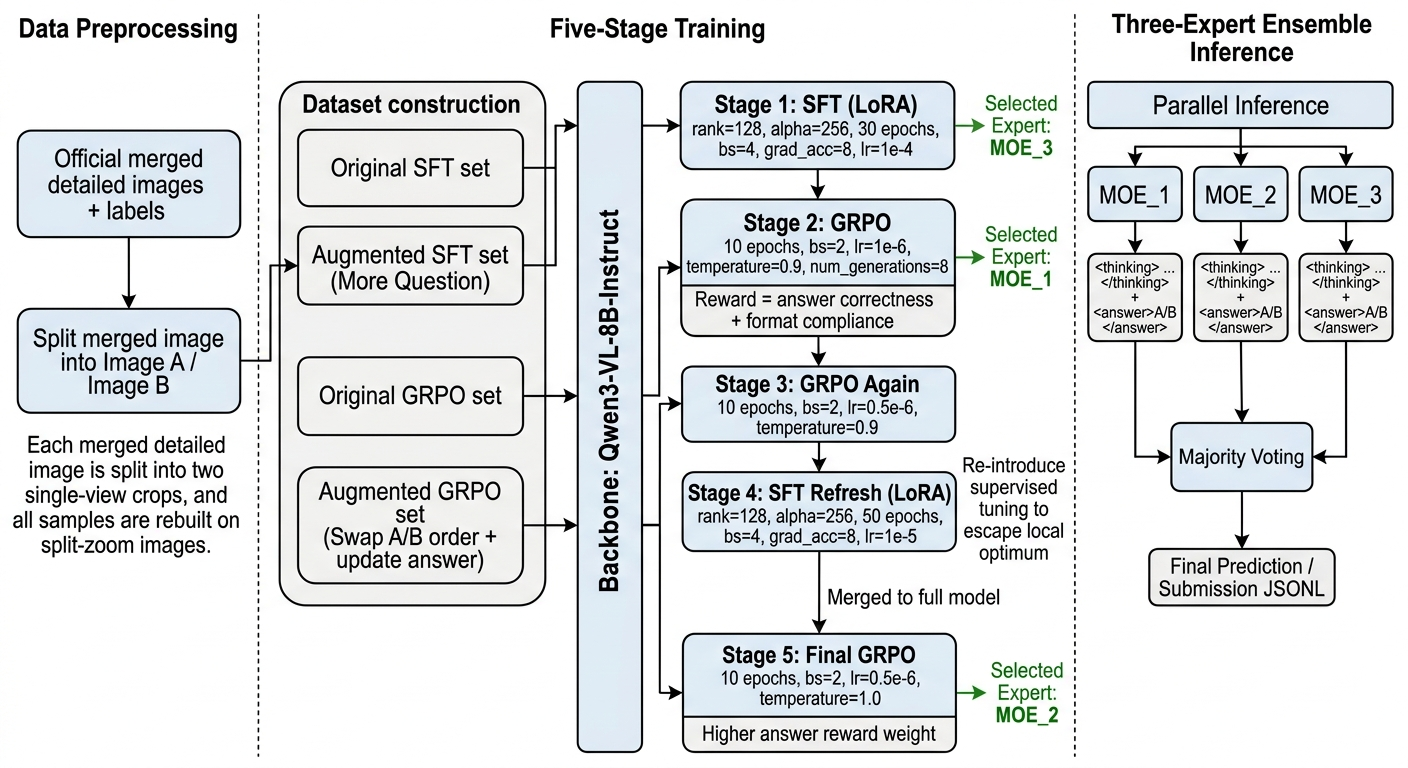}
    \caption{Overview of the proposed I$^2$ Group framework for NTIRE 2026 RAIM Track 1. Official merged detailed images are first split into paired single-view crops and reformatted into four training sets. A Qwen3-VL-8B-Instruct backbone is then progressively optimized through five stages (SFT -> GRPO -> GRPO Again -> SFT Refresh -> Final GRPO). Finally, three complementary expert checkpoints selected from Stages 1, 2, and 5 are ensembled by majority voting for robust prediction.}
    \label{fig:I2-frame}
\end{figure}

\subsubsection{Training Details}
All experiments are conducted with the \textit{ms-swift} framework on $8$ NVIDIA A800-SXM4-40GB GPUs using bfloat16 training. The visual input processing limit (\texttt{MAX\_PIXELS}) is set to $900,000$. Final inference is performed with a temperature of $0$ and \texttt{max\_new\_tokens} set to $512$. The training schedule is summarized as follows:

\noindent\textbf{Stage 1: SFT (LoRA).} Starting from \textit{Qwen3-VL-8B-Instruct}, the team performs LoRA-based supervised fine-tuning on the instruction-expanded SFT set. They use all-linear target modules with LoRA rank $r=128$, $\alpha=256$, and dropout $0.0$. The model is trained for $30$ epochs with a per-device batch size of $4$, gradient accumulation steps of $8$, learning rate of $1 \times 10^{-4}$, and maximum sequence length of $2,048$. The checkpoint from this stage is selected as the first expert model, denoted as MOE\_3.

\noindent\textbf{Stage 2: GRPO.} Based on the merged full model from Stage 1, the researchers perform full-parameter GRPO post-training on the original GRPO dataset. They employ two reward functions: answer correctness (weight $1.0$) and format compliance (weight $0.5$). Main settings include $\beta=0.04$, $8$ generations, max completion length of $512$, $10$ epochs, per-device batch size of $2$, gradient accumulation steps of $2$, learning rate of $1 \times 10^{-6}$, maximum sequence length of $4,096$, and sampling temperature of $0.9$. DeepSpeed ZeRO-3 is used for efficient training. This checkpoint is selected as MOE\_1.

\noindent\textbf{Stage 3: GRPO Again.} GRPO post-training continues from the Stage 2 checkpoint on the augmented GRPO dataset (swapped A/B order and updated answers). This stage retains the reward design of Stage 2 but reduces the learning rate to $5 \times 10^{-7}$. Other settings: $10$ epochs, per-device batch size $2$, gradient accumulation $2$, maximum sequence length $4,096$, temperature $0.9$, and $100$ maximum steps. This stage explores a better optimization region but is not selected as a final expert.

\noindent\textbf{Stage 4: SFT Refresh (LoRA).} To alleviate local optimum issues after repeated GRPO optimization, supervised tuning is re-introduced on top of the Stage 3 checkpoint. LoRA parameters are rank $r=128$, $\alpha=256$, and dropout $0.0$. Training runs for $50$ epochs with a per-device batch size of $4$, gradient accumulation of $8$, learning rate of $1 \times 10^{-5}$, and sequence length of $4,096$ using DeepSpeed ZeRO-3. LoRA weights are then merged into a full model.

\noindent\textbf{Stage 5: Final GRPO.} Using the merged model from Stage 4, a final round of full-parameter GRPO post-training is conducted on the augmented dataset. The answer reward weight is increased to $1.5$ to emphasize decision correctness (format weight remains $0.5$). Settings include $\beta=0.04$, $8$ generations, $10$ epochs, per-device batch size $2$, gradient accumulation $2$, learning rate $5 \times 10^{-7}$, sequence length $4,096$, temperature $1.0$, and $100$ maximum steps. This checkpoint is selected as MOE\_2.

During inference, MOE\_1, MOE\_2, and MOE\_3 predict independently in parallel, and the final answer is produced by majority voting across the three experts, providing superior robustness compared to any single checkpoint.

\subsubsection{Validation Performance}
The team evaluates the stage-wise models on the Phase 2 validation set, which contains $102$ image pairs. The progressive training strategy consistently improves validation accuracy, and the final three-expert ensemble achieves the best overall performance. 

\begin{table}[!ht]
\centering
\caption{Stage-wise Validation Performance of I$^2$ Group}
\label{tab:validation}
\resizebox{\columnwidth}{!}{
\begin{tabular}{lllc}
\hline
\textbf{Stage} & \textbf{Training Type} & \textbf{Validation Accuracy} & \textbf{Selected as Expert} \\
\hline
Stage 1 & SFT (LoRA) & $0.74$ ($75/102$) & Yes, MOE\_3 \\
Stage 2 & GRPO & $0.77$ ($79/102$) & Yes, MOE\_1 \\
Stage 3 & GRPO Again & $0.79$ ($81/102$) & No \\
Stage 4 & SFT Refresh (LoRA) & $0.81$ ($83/102$) & No \\
Stage 5 & Final GRPO & $0.84$ ($86/102$) & Yes, MOE\_2 \\
Ensemble & Majority Voting & $0.86$ ($88/102$) & Final System \\
\hline
\end{tabular}}
\end{table}

For the final selected ensemble, the validation submission on Phase 2 achieved an accuracy of $0.8627$ and a final score of $0.649$.

\subsection{fugui - Methodology}
\label{sec:fugui}
The fugui team proposes a dual-branch framework that separately learns the answer label and the thinking rationale. These distinct outputs are subsequently fused to enhance both the comparative accuracy and the textual quality of the reasoning process.

\subsubsection{Data Synthesis and Pre-processing} 
To augment training data, positive effects are assigned to ground-truth images to amplify the visual contrast between Image A and Image B. The team further applies global noise injection, contrast adjustments, and random rotations. To mitigate positional bias, images A and B (along with their text descriptions) are randomly swapped. Finally, four sets of localized patches are cropped from the full-resolution images to force the model's attention on fine-grained details.

\subsubsection{Network Design} 
Built upon the backbone's decoder, the architecture features two task-specific branches. Both share the same foundational backbone and LoRA fine-tuning parameters but utilize distinct input templates: \text{a) Answer Branch:} Employs a custom template optimized to directly output the binary answer, intentionally omitting redundant reasoning descriptions. \text{b) Thinking Branch:} Utilizes the standard \textit{Qwen3VL} template to explicitly guide the model in elaborating the reasoning process, thereby significantly improving interpretability scores.

\subsubsection{Training Details} 
Implemented via LLaMA-Factory, the LoRA fine-tuning (rank $r=32$, $\alpha=64$, dropout $0.1$) is applied to all layers of the base model, while the vision tower and multi-modal projector remain frozen. The network is optimized using AdamW (learning rate $5 \times 10^{-5}$, gradient clipping norm $1.0$, weight decay $0.0$) and a cosine annealing scheduler with $10$ warm-up steps. The effective batch size is $8$ (batch size $1$ with $8$ accumulation steps). Additional settings include NeFTune noise $\alpha=5$, a maximum sequence length of $16,384$. Experiments were conducted on NVIDIA A100 GPUs.

\subsection{LZ - Methodology}
\label{sec:lz}
The LZ team proposes a multi-stage training and ensemble framework built upon the \textit{Qwen3-VL-8B-Instruct} backbone. The approach relies entirely on the officially provided dataset and involves a sequential process of Supervised Fine-Tuning (SFT) and Group Relative Policy Optimization (GRPO). To ensure robust final performance, the team employs a comprehensive ensemble strategy combining predictions from multiple intermediate checkpoints during the inference phase.

\subsubsection{Data Synthesis and Pre-processing}
The researchers strictly utilize the officially provided data and explicitly avoid incorporating any external datasets. All data sets, including the original and locally processed training, validation, and test splits, are structured.

\subsubsection{Network Design and Training Details}
All training and inference processes are executed on a single node equipped with $8$ NVIDIA A100 ($80$GB) GPUs, using the pre-trained \textit{Qwen3-VL-8B-Instruct} model as the foundational weight initialization. The training pipeline is divided into two primary phases:

\begin{itemize}
    \item \textbf{Supervised Fine-Tuning (SFT):} The team initiates the pipeline with two distinct SFT tasks, resulting in two foundational model checkpoints (\text{model1} and \text{model2}).
    \item \textbf{Group Relative Policy Optimization (GRPO):} Building upon the SFT models, the researchers execute five separate GRPO training tasks with updated prompts and configurations. This stage yields a total of seven distinct model checkpoints extracted at various training intervals (e.g., checkpoints at $600$, $750$, and $900$ steps for the first GRPO task, and subsequent checkpoints at $600$, $800$, or $1,200$ steps for the remaining tasks).
\end{itemize}

\subsubsection{Inference and Ensemble Strategy}
For both the Phase $2$ validation and the final test evaluation, the team leverages a multi-checkpoint ensemble strategy to maximize decision robustness. During inference, the system independently processes the evaluation datasets through all seven saved GRPO checkpoints. The discrete outputs generated by each checkpoint are subsequently aggregated into a unified ensemble file. Finally, the team applies a custom post-processing script to the ensembled results to generate the definitive submission format.

\subsection{ongaku - Methodology}
\label{sec:ongaku}
The ongaku team's pipeline leverages the LlamaFactory framework to fine-tune a multimodal large language model (MLLM) for the NTIRE challenge. The workflow integrates custom data preprocessing involving geometric augmentations and a thinking-process masking strategy within the trainer to focus the model's learning on the final assessment results.

\subsubsection{Data Synthesis and Pre-processing}
The preprocessing stage focuses on enhancing model robustness through spatial diversity. Training images undergo a multi-stage augmentation pipeline:
\begin{itemize}
    \item \textbf{Geometric Augmentation:} The team applies standard perturbations (noise, contrast, and rotation), image swapping, and random crops to ensure the model learns localized features.
    \item \textbf{Format Conversion:} Data is structured into the ShareGPT format, mapping ``conversations'' and ``images'' to specific columns defined in a custom \texttt{dataset\_info.json} entry.
    \item \textbf{Inference Consistency:} For validation and test sets, a dedicated \texttt{random\_crop.py} script is used to ensure the image patches remain consistent with the training distribution.
\end{itemize}

\subsubsection{Network Design}
The architecture utilizes \textit{Qwen3-VL} as the backbone, integrated with LlamaFactory. A critical modification was implemented in the core training logic to optimize the learning signal:
\begin{itemize}
    \item \textbf{Thinking-Masked Loss Computation:} The researchers modified the trainer to implement a customized loss calculation.
    \item \textbf{Selective Backpropagation:} Specifically, the ``thinking'' tokens (the chain-of-thought or intermediate reasoning steps) are masked out during the loss computation. This ensures that the gradient updates are mostly driven by the accuracy of the final output, preventing the model from being penalized for stylistic variations in its internal reasoning process while enforcing high-quality final results.
\end{itemize}

\subsubsection{Training Details}
The framework is developed using PyTorch and the LlamaFactory toolkit, centered on \textit{Qwen3-VL-8B-Instruct} as the multimodal backbone. To balance performance and efficiency, the team employs 4-bit quantization via \texttt{bitsandbytes} with double quantization enabled. The fine-tuning process utilizes the LoRA strategy targeting all linear layers.

Training is conducted on a server equipped with dual NVIDIA A100 (40GB) GPUs using BF16 mixed-precision. To stabilize the training trajectory, the researchers set a global gradient clipping threshold of 1.0 and implement a gradient accumulation strategy to achieve an effective batch size of 8. Furthermore, they incorporate NeFTune noise to enhance model robustness, with an input sequence cutoff of 16,384 tokens and a dynamic image resolution range spanning approximately 1K to 3.5M pixels.

\subsection{test123djak - Methodology}
\label{sec:test123djak}
The test123djak team proposes a dual-model framework based on Qwen3-VL-8B-Instruct, independently training an Answer Model dedicated to final predictions and a Thinking Model dedicated to rationale generation. To efficiently adapt the shared foundation backbone to both tasks, Low-Rank Adaptation is applied to all linear layers. Notably, both the vision model and the multi-modal aligner remain unfrozen during training to optimally capture the specific visual distribution of the challenge data.

\subsubsection{Data Pre-processing} 
As a foundational pre-processing step, the concatenated input images are explicitly split into individual frames and processed at their original high resolutions (up to $5,000,000$ pixels) to strictly preserve fine-grained visual details. The dataset is further augmented via metadata swapping and horizontal flipping to increase diversity. Crucially, this augmented dataset is partitioned into two independent subsets, with one isolating the final comparative answers and the other isolating the reasoning processes—to separately supervise the two specialized models.

\subsubsection{Training and Inference Details} 
Both models undergo Supervised Fine-Tuning (SFT) for $30$ epochs with DeepSpeed Zero-2 optimization, BF16 precision, and Flash Attention 2. The training configuration employs a learning rate of $1 \times 10^{-4}$, a per-device batch size of $8$, $12$ gradient accumulation steps, and a $0.05$ warmup ratio, with gradient checkpointing enabled for memory efficiency. For deterministic inference, the fine-tuned LoRA weights are merged into the base model. The sequential generation of the thinking texts and final answers is then accelerated using vLLM, configured with a tensor parallel size of $2$ and a temperature of $0.0$. All experiments were conducted on two NVIDIA PRO 6000 GPUs.

\subsection{dr0strange - Methodology}
\label{sec:dr0strange}
The dr0strange team proposes a two-stage pipeline for image quality comparison and explanation generation. Each sample contains two concatenated images: one global view and one detail crop. In each concatenated image, candidate A is on the left and candidate B is on the right. The team first splits both concatenated images into individual A/B images and extracts six complementary no-reference image quality assessment (NR-IQA) metrics: BRISQUE, NIQE, PIQE, NRQM, CLIPIQA, and HyperIQA. These features capture both traditional statistical distortions and deep perceptual quality cues. A LightGBM (LGBM) classifier is then used to predict which image has better perceptual quality based on the aggregated feature representation. In the second stage, the researchers leverage a fine-tuned multimodal large language model (MLLM) to generate natural language explanations for the predicted preference. By conditioning on both the visual inputs and the model decision, the MLLM produces interpretable reasoning that aligns with human perception, such as differences in sharpness, noise, and contrast.

\subsubsection{Data Synthesis and Pre-processing}
Each input sample consists of two concatenated images: one global view and one detail crop. In each image, candidate A appears on the left and candidate B appears on the right. The team splits both concatenated images into individual A/B images according to their spatial arrangement. For each image, the researchers compute six NR-IQA metrics:
\begin{itemize}
    \item \textbf{BRISQUE, NIQE, PIQE:} Traditional statistical features based on natural scene statistics.
    \item \textbf{NRQM:} A learning-based quality regression metric.
    \item \textbf{CLIPIQA, HyperIQA:} Deep feature-based perceptual quality metrics.
\end{itemize}
These metrics provide a compact yet expressive representation of image quality from multiple perspectives. The resulting feature vectors for image A and image B are concatenated to form the final input to the classifier.

\subsubsection{Network Design}
The architecture consists of two primary modules:

\noindent\textbf{Quality Comparison Module (LightGBM):} The team adopts a LightGBM (LGBM) classifier to predict the relative perceptual quality between image A and image B. The input is the concatenated IQA feature vector, and the output is a binary label indicating the preferred image. LightGBM is chosen for its strong ability to model non-linear relationships across heterogeneous features, as well as its efficiency and robustness on tabular data. This makes it particularly suitable for aggregating diverse IQA metrics into a unified decision.

\noindent\textbf{Explanation Generation Module (MLLM):} A fine-tuned multimodal large language model is used to generate explanations for the predicted preference. Given the image pair and the classifier output, the MLLM takes visual inputs along with a structured prompt to produce natural language reasoning. The generated explanations focus on perceptual factors such as sharpness, noise level, texture fidelity, and contrast, ensuring that the reasoning is consistent with both the visual evidence and the IQA-based decision.

\subsubsection{Training Details}
The IQA feature extractors are pre-trained and used directly without additional training. The LightGBM classifier is trained on labeled image pairs using supervised learning to optimize classification accuracy. Hyperparameters such as the number of leaves, learning rate, and maximum depth are tuned to achieve optimal performance.

For the explanation generation module, the MLLM is fine-tuned using supervised fine-tuning (SFT) on datasets containing image pairs with annotated preference explanations. The team uses \textit{Qwen3-VL-8B-Thinking} as the backbone model. The training pipeline is implemented with \textit{ms-swift} on top of PyTorch and is executed on multi-GPU hardware. In practice, the researchers use LoRA-based adaptation together with DeepSpeed ZeRO-3, \text{bfloat16} training, and FlashAttention for memory-efficient and stable optimization under high-resolution visual token settings.

\subsection{NTR - Methodology}
\label{sec:ntr}
The NTR team proposes a two-stage fine-tuning pipeline for Qwen3-VL-8B-Instruct, seamlessly combining Supervised Fine-Tuning (SFT) and Group Relative Policy Optimization (GRPO). Utilizing Low-Rank Adaptation, the model processes images simultaneously to generate a chain-of-thought rationale followed by a binary preference label. 

\subsubsection{Data Pre-processing} 
To ensure optimal visual perception, training images are resized to a maximum side length of 1536 pixels. Furthermore, to explicitly mitigate positional bias where the model might arbitrarily favor the first image, the original 100 image pairs are augmented by swapping Image A and Image B while simultaneously flipping their corresponding labels. This strategy effectively doubles the training dataset to 200 samples.

\subsubsection{Training Details} 
During the initial stage, the backbone undergoes SFT on the augmented dataset for 5 epochs using the AdamW optimizer, establishing a warm-start checkpoint. In the subsequent stage, the team executes three independent GRPO runs using custom IQA-specific reward functions. The baseline GRPO run optimizes primarily for binary accuracy and format compliance. The subsequent runs build upon this by introducing additional nuanced rewards for IQA attribute coverage and comparative language quality. All GRPO configurations utilize 8 rollout generations per sample to ensure robust policy updates. 

\subsubsection{Ensemble Strategy} 
For the final inference phase, the team employs a majority voting ensemble across four diverse checkpoints derived from the GRPO runs. In the event of a tie, a designated primary checkpoint serves as the definitive tiebreaker. To maintain logical consistency between the text and the final choice, the submitted reasoning rationale is extracted directly from the first checkpoint that aligns with the majority answer. Training and inference were accelerated using NVIDIA A100 and H100 GPUs.

\end{document}